\documentclass[conference]{IEEEtran}
\IEEEoverridecommandlockouts
\usepackage{cite}
\usepackage{amsmath,amssymb,amsfonts}
\usepackage{algorithmic}
\usepackage{graphicx}
\usepackage{textcomp}
\usepackage{xcolor}
\usepackage{url}
\usepackage{algorithm}
\usepackage{hyperref}
\usepackage{orcidlink}
\usepackage[a4paper, total={184mm,239mm}]{geometry}
\def\BibTeX{{\rm B\kern-.05em{\sc i\kern-.025em b}\kern-.08em
    T\kern-.1667em\lower.7ex\hbox{E}\kern-.125emX}}
\begin{document}

\title{Attention Consistency Regularization for Interpretable Early-Exit Neural Networks\\
}

\author{\IEEEauthorblockN{Yanhua Zhao
}}
\maketitle

\begin{abstract}
Early-exit neural networks enable adaptive inference by allowing predictions at intermediate layers, reducing computational cost. However, early exits often lack interpretability and may focus on different features than deeper layers, limiting trust and explainability. This paper presents Explanation-Guided Training (EGT), a multi-objective framework that improves interpretability and consistency in early-exit networks through attention-based regularization. EGT introduces an attention consistency loss that aligns early-exit attention maps with the final exit. The framework jointly optimizes classification accuracy and attention consistency through a weighted combination of losses. Experiments on a real-world image classification dataset demonstrate that EGT achieves up to 98.97\% overall accuracy (matching baseline performance) with a 1.97× inference speedup through early exits, while improving attention consistency by up to 18.5\% compared to baseline models. The proposed method provides more interpretable and consistent explanations across all exit points, making early-exit networks more suitable for explainable AI applications in resource-constrained environments.
\end{abstract}

\begin{IEEEkeywords}
Early exit networks, explainable AI, attention mechanisms, multi-objective learning.\end{IEEEkeywords}

\section{Introduction}

Deep neural networks have achieved strong performance across computer vision tasks, but their computational cost limits deployment in resource-constrained settings. Early-exit networks address this by enabling predictions at intermediate layers, allowing simple samples to exit early and reducing average inference time \cite{b1}. However, these networks face a challenge: early exits may focus on different features than deeper layers, reducing explanation consistency.

Explainable AI (XAI) is increasingly important for trust and understanding, especially in critical applications \cite{b2}. Attention mechanisms provide spatial explanations by highlighting important regions, but without explicit regularization, early exits can produce inconsistent attention patterns. This inconsistency undermines trust and makes it difficult to understand what the network learns at different depths.

This paper introduces Explanation-Guided Training (EGT), a training framework that improves interpretability and consistency in early-exit networks. EGT uses a multi-objective loss that combines: (1) classification loss for accuracy, and (2) attention consistency loss to align early-exit attention with the final exit. By regularizing attention during training, EGT ensures that early exits learn consistent and interpretable explanations while maintaining competitive accuracy.

\section{Methods}

\subsection{Model Architecture}
%As shown in Figure~\ref{fig:model_structure},
The proposed framework employs an Early-Exit Convolutional Neural Network (CNN) with integrated attention mechanisms. The architecture consists of five convolutional layers with progressively increasing feature dimensions: 64, 128, 256, 512, and 512 channels, respectively. Each convolutional layer is followed by batch normalization, ReLU activation, and max pooling operations, with the final layer utilizing adaptive average pooling for global feature aggregation.

The network incorporates five early exit points, where each exit point $i \in \{1, 2, 3, 4, 5\}$ is equipped with a dedicated attention module and a classification head. The attention module at each exit generates spatial attention maps that highlight salient regions in the feature representations, thereby providing interpretability while guiding the learning process.

\begin{figure}[b] % [H] forces placement exactly here
    \centering
    \includegraphics[width=0.3\textwidth]{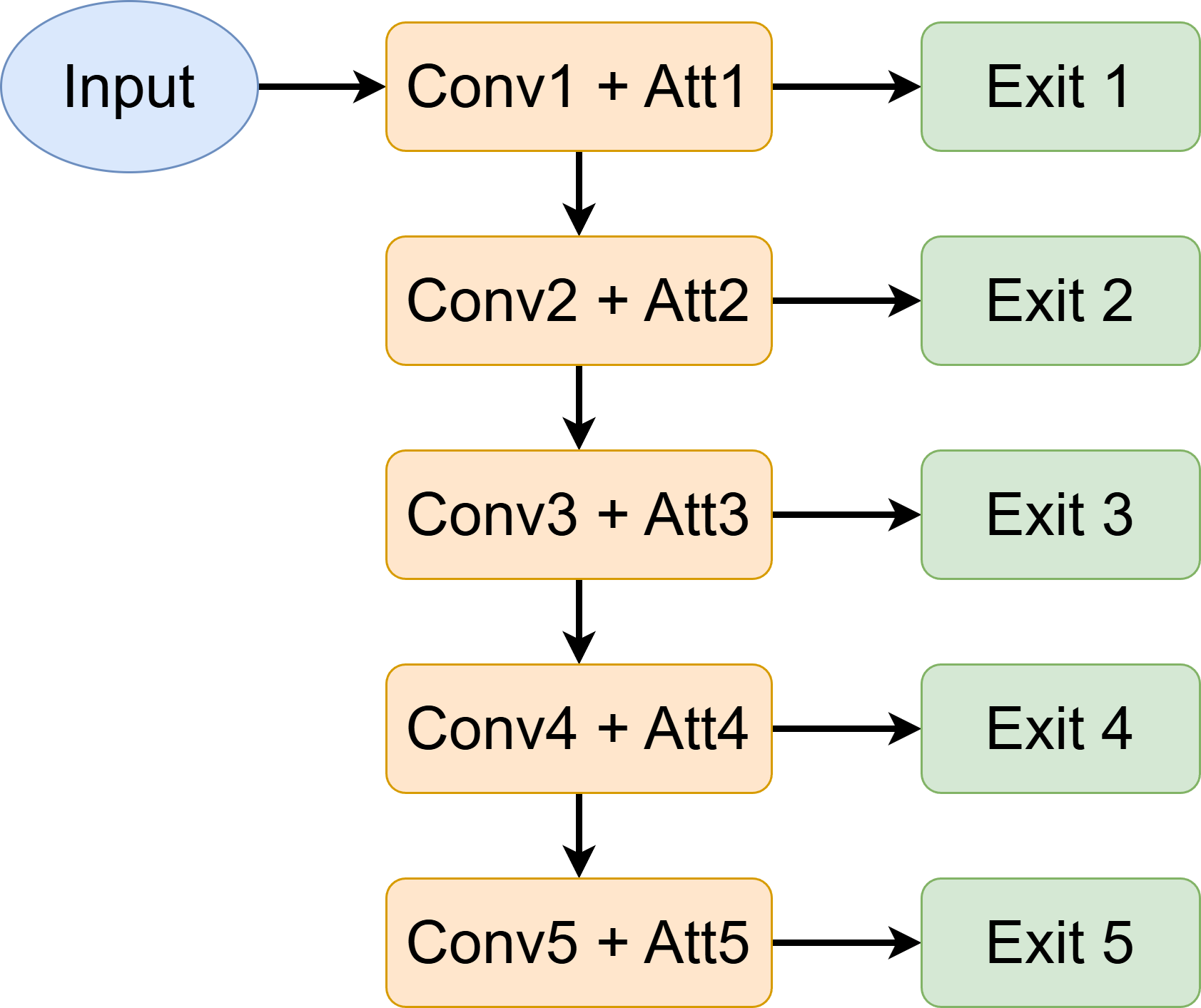} 
    \caption{Overview of the EGT model architecture, showing the early-exit branches and attention mechanisms.}
    \label{fig:model_structure}
\end{figure}

\subsection{Explanation-Guided Training Framework}
The Explanation-Guided Training (EGT) framework introduces a multi-objective loss function that jointly optimizes classification accuracy and attention map quality. The total loss function $\mathcal{L}_{\text{total}}$ is formulated as:
$$\mathcal{L}_{\text{total}} = \mathcal{L}_{\text{cls}} + \alpha \cdot \mathcal{L}_{\text{consistency}}$$
where $\alpha$ is hyperparameters controlling the relative importance of attention consistency loss.

The attention consistency loss $\mathcal{L}_{\text{consistency}}$ enforces similarity between attention maps from early exits and the final exit, ensuring that early exits learn to focus on the same discriminative regions as the deeper network. For each early exit $i \in \{1, 2, 3, 4\}$, the attention map $A_i$ is compared with the final attention map $A_5$ using cosine similarity:
$$
\mathcal{L}_{\text{consistency}}
= \frac{1}{4} \sum_{i=1}^{4} d_{\cos}(\tilde{A}_i, A_5),
$$
$$ 
d_{\cos}(x,y) = 1 - \frac{x^\top y}{\|x\|\|y\|}.
$$
where $\tilde{A}i$ denotes the bilinearly interpolated version of $A_i$ to match the spatial dimensions of $A_5$.

\section{Experimental Evaluation}
\subsection{Experimental Setup}
Experiments were conducted on a real-world image classification dataset \cite{b7} comprising 9 classes, with 1,363 training samples and 1,364 test samples. The proposed EGT model was trained for 50 epochs using the Adam optimizer with an initial learning rate of 0.001, which was decayed by a factor of 0.5 every 15 epochs using a step-wise learning rate scheduler. 

The baseline model employed an identical architecture and training configuration but without the explanation-guided regularization terms. Early exit decisions were made using a confidence threshold of 0.9, where samples exiting at confidence levels below this threshold proceeded to subsequent exits.

The EGT framework introduces two key hyperparameters. The coefficient $\alpha$ controls the contribution of the attention consistency loss; in our experiments, we varied $\alpha$ over the range from $0.1$ to $0.5$ in incremental steps.

\subsection{Attention Consistency Analysis}
\renewcommand{\arraystretch}{1.3}
\begin{table}[h]
    \centering
    \caption{EGT Models Attention Consistency.}
    \label{tab:attention_consistency}
    \begin{tabular}{p{1.8cm} p{0.7cm} p{0.7cm} p{0.7cm} p{0.7cm} p{0.4cm} p{1.1cm}}
        \hline\hline
        Model & Exit 1 & Exit 2 & Exit 3 & Exit 4 & Avg & Overall Acc. (\%)\\
        \hline
        Baseline Model & 0.769 & 0.736 & 0.786 & 0.482 & \textbf{0.693} & 98.97\\
        \hline
        EGT ($\alpha$ = 0.1) & 0.797 & 0.817 & 0.819 & 0.814 & 0.812 &  97.73\\
        EGT ($\alpha$ = 0.2) & 0.751 & 0.787 & 0.79 & 0.786 & 0.778 &98.39 \\
        EGT ($\alpha$ = 0.3) & 0.802 & 0.826 & 0.829 & 0.828 & \textbf{0.821} &98.46 \\
        EGT ($\alpha$ = 0.4) & 0.79 & 0.838 & 0.844 & 0.778 & 0.813 &98.97 \\
        EGT ($\alpha$ = 0.5) & 0.7 & 0.793 & 0.795 & 0.783 & 0.768 & 97.8\\
        \hline\hline
    \end{tabular}
    \label{tab1}
\end{table}

%%%%%%%%%%%%%%%%%%%%%%%%%%%

Table \ref{tab1} presents attention consistency results. The baseline achieves 0.693 average consistency. All EGT configurations improve consistency, ranging from 0.768 ($\alpha$ = 0.5) to 0.821 ($\alpha$ = 0.3), representing 10.8\% to 18.5\% improvements over baseline. Exit 4 shows the largest improvement, increasing from 0.482 (baseline) to values between 0.778 and 0.828 across EGT models, representing relative improvements of 61.4\% to 71.8\%.
Classification accuracy remains competitive across all EGT models (97.73\%–98.97\%). Notably, EGT ($\alpha$ = 0.4) achieves identical accuracy to the baseline (98.97\%) while improving consistency by 17.3\% (0.813 vs 0.693). EGT ($\alpha$ = 0.3) achieves the highest consistency (0.821, 18.5\% improvement) with 98.46\% accuracy, demonstrating effective balance between interpretability and performance. The configuration with $\alpha$ = 0.5 exhibits lower consistency (0.768) compared to intermediate $\alpha$ values, suggesting an optimal range for the consistency regularization weight.

\subsection{Inference Efficiency Analysis}
Table \ref{tab2} reports the average runtime per sample for the early-exit model versus the model without early exit (always using Exit 5), showing results only for $\alpha = 0.5$. The early-exit model achieves an average runtime of 1.83 ms per sample, compared to 3.6 ms for the model without early exit, corresponding to a 1.97× inference-time speedup. This highlights the efficiency gained by incorporating early exits into the EGT framework.
\renewcommand{\arraystretch}{1.3}
\begin{table}[h]
    \centering
    \caption{EGT Models Inference Efficiency.}
    \label{tab:attention_consistency}
    \begin{tabular}{lcc}
        \hline\hline
        Model & Avg Time/Sample (ms) & Accuracy (\%)\\
        \hline
        With Early Exit & 1.83 & 97.8\\
        Without Early Exit & 3.6 & 99.56\\
        \hline\hline
    \end{tabular}
    \label{tab2}
\end{table}

\section{Conclusion and Future Work}
\label{sec:conclusion}
This paper presents Explanation-Guided Training (EGT), a multi-objective framework that improves interpretability in early-exit neural networks through attention-based regularization. All EGT configurations with varying $\alpha$ values (0.1 to 0.5) improve attention consistency compared to the baseline. EGT ($\alpha$ = 0.3) achieves the highest average consistency of 0.821, representing an 18.5\% improvement over the baseline (0.693). Exit 4 shows the largest improvement, with consistency increasing from 0.482 to 0.778–0.828 (61.4\% to 71.8\% relative improvement). EGT models maintain competitive accuracy (97.73\% to 98.97\%), with EGT ($\alpha$ = 0.4) matching baseline accuracy (98.97\%) while improving consistency by 17.3\%. The early-exit mechanism provides a 1.97× speedup (1.83 ms vs 3.6 ms per sample) while maintaining 97.8\% accuracy, demonstrating that EGT successfully maintains computational efficiency while enhancing interpretability.

%%%%%%%%%%%%%%%%%%%%%%%%%%%

Future work will explore several directions, including integrating the EGT framework with other model architectures, testing on larger and  more diverse datasets, developing adaptive regularization weights for exit-specific optimization, enforcing hierarchical consistency between exits, providing theoretical analysis of consistency bounds and optimizing the framework for real-time edge deployment.

%\section*{Acknowledgment}

%The preferred spelling of the word ``acknowledgment'' in America is without 
%an ``e'' after the ``g''. Avoid the stilted expression ``one of us (R. B. 
%G.) thanks $\ldots$''. Instead, try ``R. B. G. thanks$\ldots$''. Put sponsor 
%acknowledgments in the unnumbered footnote on the first page.

%\section*{References}

\vspace{12pt}

\end{document}